\documentclass{article}
\usepackage{ijcai19}

\usepackage{times}
\usepackage{soul}
\usepackage{url}
\usepackage[draft]{hyperref}
\usepackage[utf8]{inputenc}
\usepackage[small]{caption}
\usepackage{graphicx}
\usepackage{amsmath}
\usepackage{booktabs}
\usepackage{color}
\usepackage{multirow}
\usepackage{caption}
\usepackage{titlesec}
\usepackage{setspace}

\title{An Interactive Insight Identification and Annotation Framework \\ for Power Grid Pixel Maps using DenseU-Hierarchical VAE}

\author{
Tianye Zhang$^1$\footnote{Contact Author}\and
Haozhe Feng$^1$\and
Zexian Chen$^{1}$\And
Can Wang$^{1}$\And
Yanhao Huang$^{2}$\And
Yong Tang$^{2}$\And
Wei Chen$^1$\\
\affiliations
$^1$Zhejiang University\\
$^2$China Electric Power Research Institute\\
\emails
\{zhangtianye1026,fenghz,zexianchen,wcan\}@zju.edu.cn,
\{hyhao,tangyong\}@epri.sgcc.com.cn,
chenwei@cad.zju.edu.cn
}

\begin{document}
\renewcommand{\baselinestretch}{0.9}

\maketitle

\begin{abstract}
Insights in power grid pixel maps (PGPMs) refer to important facility operating states and unexpected changes in the power grid. Identifying insights helps analysts understand the collaboration of various parts of the grid so that preventive and correct operations can be taken to avoid potential accidents. Existing solutions for identifying insights in PGPMs are performed manually, which may be laborious and expertise-dependent. In this paper, we propose an interactive insight identification and annotation framework by leveraging an enhanced variational autoencoder (VAE). In particular, a new architecture, DenseU-Hierarchical VAE (DUHiV), is designed to learn representations from large-sized PGPMs, which achieves a significantly tighter evidence lower bound (ELBO) than existing Hierarchical VAEs with a Multilayer Perceptron architecture. Our approach supports modulating the derived representations in an interactive visual interface, discover potential insights and create multi-label annotations. Evaluations using real-world PGPMs datasets show that our framework outperforms the baseline models in identifying and annotating insights.
\end{abstract}

\begin{spacing}{0.95} 
\section{Introduction}
%background, motivation, challenge, existing models, our method, contribution 
%电网仿真数据分析中，pattern mining是重要的，包括哪些pattern。
%为什么通过可视化图标挖掘pattern：1.人对图像比对数字敏感；2.有时无法获取数据，只有图表

%第一段。提出问题，对一个没有label数据的检测分析的问题。

Transient stability analysis (TSA) based on power grid (PG) simulation data is one of the most challenging problems in PG operation~\cite{yan2015cascading}. Analysts need to take preventive and correct controls in the grid based on insights identified from the operating states in TSA, so that serious failures and blackouts can be avoided. Existing TSA insight identification methods require expertise and are performed manually, making it a laborious process. Generally, insights are interesting facts that can be derived from data~\cite{lin2018bigin4}. 
%They can be quite simple to be identified by a single query operation in the data, and can also be complicate to be captured by multiple steps of queries. 
For example, a TSA insight can be `\textit{power grid changes from a stable state to an unstable state where the voltages of nodes keep increasing}'. 
%\textbf{Example 2:} a fault occurs at time \textit{t} and leads to the collapse of the grid.

Considering that humans are more sensitive to visual representations of data, visualization-based approaches have been proposed to generate chart images from numerical PG simulation data to convey quantitative information~\cite{wong2009novel}. Though effective, this scheme encounters three challenging problems. First, humans can possibly deal with hundreds of charts but can hardly analyze a dataset containing thousands of charts. Second, sometimes analysts can only access chart representations without the underlying data, for example, report documents or simulation tools. Third, the diversity of TSA insights makes it difficult to obtain a labeled dataset. Therefore, a proper solution for TSA insight identification in unlabeled PG chart images is demanded. As a first attempt for this purpose, we focus on one particular kind of charts, power grid pixel maps (PGPMs) variations in PG bus variables including voltage, frequency, rotor angle etc.. Buses in a PG are important facilities used to dispatch and transmit electricity. Bus variables are important indicators for PG running states and serve as crucial factors for PG operation and control decisions.
%as it is one of the most commonly used types in TSA.

% existing methods and challenges
% supervised: lack of labels
% unsupervised: 1.multi-label clustering? 2.work on natural image
%第二段。目前有哪些方法，他们的缺点是啥（重点）。

%The literature has shown abundant image pattern mining models~\cite{boutell2002review}, including supervised models represented by convolutional neural networks (CNN)~\cite{krizhevsky2012imagenet} and unsupervised models like clustering~\cite{jain1988algorithms,xie2016unsupervised}. Although these models have show impressing performance on natural scenes, they are barely studied on chart images, which is usually perceived in a different way~\cite{cleveland1984graphical}. 

%Chart images are usually perceived in a different way than natural scenes. 
Existing chart recognition studies employ supervised models and focus on classifying chart types and decoding visual contents~\cite{savva2011revision}. Models like support vector machine (SVM) and convolutional neural networks (CNN) have been widely used for this purpose. 
%Despite the impressing performance they achieved, they have barely been tested on large data sets containing more than 10,000 charts.
A fundamental limitation of this strategy is that only shape features are exploited. Some state-of-the-art methods such as visual question answering models utilize additional semantic features, but may suffer from low accuracy in answering chart image questions~\cite{kafle2018dvqa}. Nevertheless, few studies have been attempted to learn strong features using unsupervised models, due to the following two unresolved problems: (i) How to define a TSA insight and describe it by semantic features? (ii) How to annotate a dataset with identified TSA insights when each PGPM may contain a combination of insights?

%our methods
%contributions：1.visualization driven approach; 2.application domain
%第三段，我们提出技术123，解决缺点123。 然后说一下整体流程什么的

\begin{figure}[!tbp]
\centering
\setlength{\abovecaptionskip}{3pt}
\includegraphics[width = 0.9 \columnwidth]{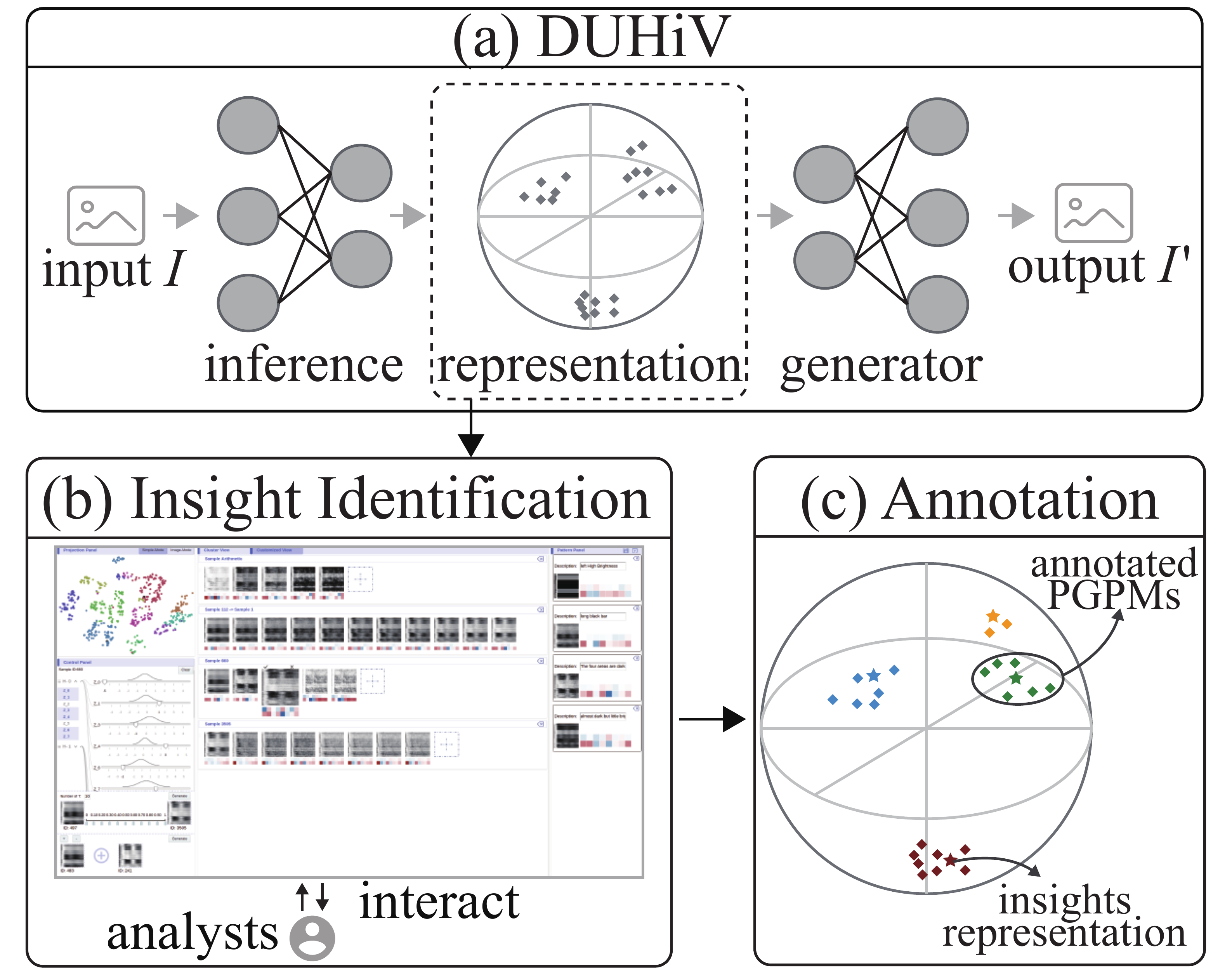}
\caption{The pipeline of our method. (a) A DenseU-Hierarchical VAE (DUHiV) model used to learn the representation of PGPMs. (b) A visual interface used to visualize the learned representations, identify insights and generate representations for insights. (c) An annotation module that compares the representations of PGPMs and insights.}
\label{pipeline}
\vspace{-5mm}
\end{figure}

Our solution for these problems is 
%an interactive analysis scheme that assists PG experts effectively define TSA insights based on arithmetic operations on representations and perceptual level visual depictions. In this way, each PGPM can be annotated with multiple insights by analyzing possible insights and the PGPM's local structures in the same latent space. To effectively capture important information in PGPMs, we propose 
a novel interactive insight identification and annotation framework for unlabeled PGPMs (Figure~\ref{pipeline}). The process starts by training a DenseU-Hierarchical VAE (DUHiV) to learn the representations of PGPMs (Figure~\ref{pipeline} (a), Section~\ref{mmodel}). Analysts are then able to perform arithmetic operations on the derived representations and apply visualization-based approaches to analyze the semantic changes in the generated PGPM brought by such operations. In this way, analysts can interactively identify insights and generate the representation for each insight (Figure~\ref{pipeline} (b), Section~\ref{minterface}). Finally, the dataset can be annotated with the identified TSA insights by comparing the representation of insights and the PGPMs.(Figure~\ref{pipeline} (c), Section~\ref{mannotation}). 
%For newly acquired data, TSA insights can also be identified once its representation is obtained from the trained DUHiV model.

This paper presents the following contributions:
\begin{itemize}
\vspace{-1.5mm}
\setlength{\itemsep}{0pt}
\setlength{\parsep}{0pt}
\setlength{\parskip}{0pt}
%    \item 
%    We propose an interactive end-to-end framework which advances the real-world power grid transient stability analysis (TSA) problem.
    
%    \item
%    We introduce novel visual representations of the output of DUHiV for effectively discovering distinctive insights. 
    %interactive visualization techniques to interpret the representations learned from DUHiV and perform arithmetic operations to generate the ultimate representations, enabling more effective insight identification and more accurate annotation.
%    \item
%    We collect PGPMs of real-world PG simulation data to evaluate our method. Experimental results demonstrate significant performance improvement of our method over baselines.
    \item 
    We design a new hierarchical VAE architecture, DUHiV, that achieves a significantly tighter evidence lower bound (ELBO).
    \item
    We develop a visual interface that efficiently supports interactive insight identification by conducting arithmetic operations on the learned representations.
    \item
    We propose effective unsupervised and semi-supervised multi-label annotation methods used for annotating the dataset with identified insights.
\end{itemize}

\section{Problem Definition}
%用符号化的描述

%We first provide the preliminary concept of a TSA pattern, and then we state the problem of TSA pattern identification in unlabeled PGPMs.

%数据是什么：pixel map
Let $\mathcal{G}=\{v_1,v_2,...,v_M\}$ be a bus set of size $M$. The set of PGPMs is denoted as $\mathcal{I}=\{I_1,I_2,...,I_N\}$, where each $I_i$ represents an PGPM, showing the voltage variation of the buses $\mathcal{G}$ in a period of time. Figure~\ref{vaeimage} (a) provides an example set $\mathcal{I}$ of PGPMs. The horizontal axis of each PGPM represents time and the vertical axis represents buses $\mathcal{G}$, sorted by their ID. The \textit{voltage} values are depicted by the grayscale of pixels. We focus our study on variable \textit{voltage} because it is one of the most significant and affected variables in TSA problems~\cite{kundur2004definition}. In fact, our work can be extended to PGPMs of other variables like frequency and rotor angle.

A TSA insight refers to certain interesting facts extracted from a PGPM $I_i$ and is defined as follows:

%preliminary concepts: pattern的定义: 一段时间、哪些设备、变化情况
\noindent \textbf{Definition (TSA insight)}
\textit{A TSA insight is the variable variations in a subset $\widehat{\mathcal{G}}$ of $\mathcal{G}$ from a start time to an end time. Such variations can be periodic, correlated and anomalous, etc..}

%问题是什么，目标是什么
%最终objective：identify一个pattern set，用这个pattern set label image set
The objective is to (i) identify a TSA insight set $A=\{a_1,a_2,...,a_K\}$ in $\mathcal{I}$ and (ii) perform multi-label annotation on the dataset by constructing a vector $Y_i=\{a^i_1,a^i_2,...,a^i_K\}$ for each $I_i$ where $a^i_j \in Y_i$ is assigned as 1 if insight $j$ is identified in $I_i$ and 0 otherwise, denoted as $R=(I_i,Y_i)^N_{i=1}$.

\end{spacing}

\section{Methodology}
%1.用VAE学习PGPM的表达
\subsection{DenseU-Hierarchical VAE (DUHiV)}
\label{mmodel}

\begin{figure}[!htbp]
\centering
\setlength{\abovecaptionskip}{3pt}
\includegraphics[width = 0.85 \columnwidth]{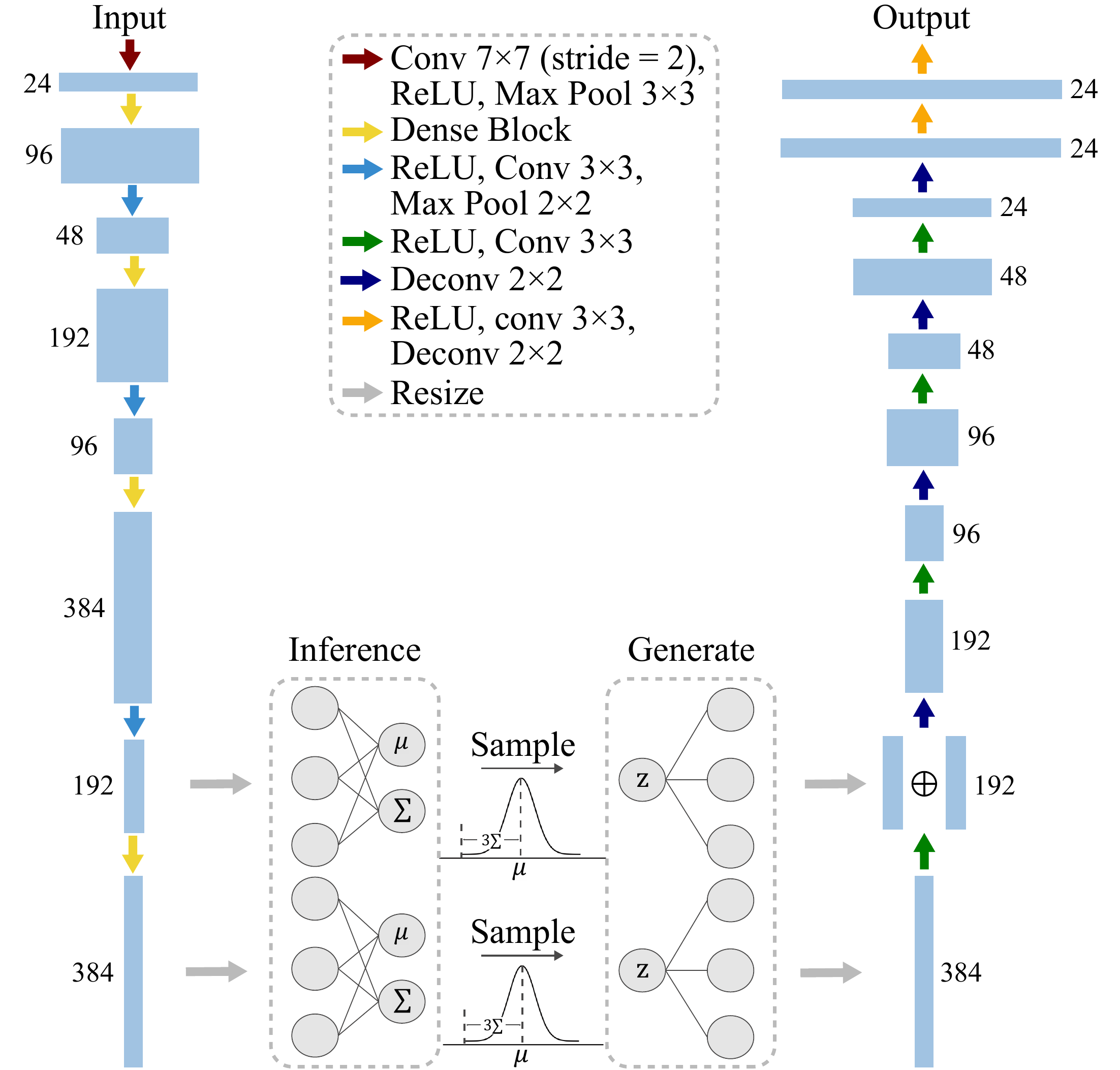}
\caption{The DUHiV architecture with two layers of latent variables. The dense block is used in the inference process (left) and the U-Net is used in the generative process (right). The posterior distribution of the hierarchical latent variables are inferenced by fully connection layers (middle).}
\label{duhiv}
\vspace{-5mm}
\end{figure}

We design DenseU-Hierarchical VAE (DUHiV) to learn the representation of PGPMs. Following the architecture presented in~\cite{DBLP:conf/icml/RezendeMW14}, we achieve further improvement by using dense block~\cite{DBLP:conf/cvpr/HuangLMW17} in the inference process and a symmetric U-Net expanding path~\cite{DBLP:conf/miccai/RonnebergerFB15} in the generative process to capture the hierarchical structure of latent variables and accommodate the large size of PGPMs. We design two hierarchies consisting of 2 layers and 4 layers of latent variables respectively for two datasets of different sizes. (Section~\ref{dataset}). The network architecture of the 2-layer DUHiV is presented in Figure~\ref{duhiv}.

%Similar to traditional hierarchical VAE models,  whose dimension is much less than the original high-dimensional PGPM set $\mathcal{I}$. For each $I \in \mathcal{I}$, there exists at least one setting of latent variables $z \in Z$ to assure the model generates something very similar to $I$. Meanwhile, we assume $X$ and $z$ are both random variables that satisfy a certain prior distribution. We adopt a typical hierarchical framework consisting an encoder and a decoder. Given an input $I$, the encoder learns the posterior distribution $p(z|I)$ of the latent space and the decoder achieves reconstruction from the latent variables $p(I|z)$.  

DUHiV assumes that a given PGPM, denoted as $\mathbf{x}$, can be represented by a set of latent variables $\mathbf{z}$, from which something very similar to $\mathbf{x}$ can be reconstructed. Therefore, DUHiV simultaneously trains an inference model, which learns the posterior distribution $q_{\phi}(\mathbf{z}\vert \mathbf{x})$ of latent variables, and a generative model $p_{\theta}(\mathbf{x},\mathbf{z})$$=p_{\theta}(\mathbf{x}\vert \mathbf{z})p_{\theta}(\mathbf{z})$, which reconstructs $\mathbf{x}$, by maximizing the likelihood:
\vspace{-2mm}
\begin{gather}
\label{likelihood}
    p_{\theta}(\mathbf{x})=\int_z p_{\theta}(\mathbf{x},\mathbf{z})dz,
\end{gather}
\noindent as well as minimizing the Kullback-Leibler divergence between the inferenced posterior distribution $q_{\phi}(\mathbf{z} \vert \mathbf{x})$ and the groundtruth distribution $p(\mathbf{z}\vert \mathbf{x})$:
\begin{gather}
\label{kl}
    KL(q_{\phi}(\mathbf{z} \vert \mathbf{x})\Vert p(\mathbf{z} \vert \mathbf{x})).
\end{gather}

In the generative model $p_\theta$, the latent variables $\mathbf{z}$ are split into $L$ layers $\mathbf{z}_i, i=1,\ldots L$ , and the generative process is described as follows:
\[
\vspace{-2mm}
\begin{split}
\vspace{-2mm}
p(\mathbf{z}) &= \Pi_{i=1}^{L}p(\mathbf{z}_i), p(\mathbf{z}_i)  \mathcal{N}(0,I),i\in 1,\ldots,L \\
q_{\phi}(\mathbf{z} \vert \mathbf{x}) &= \Pi_{i=1}^{L} q_{\phi}(\mathbf{z}_i \vert \mathbf{x}), q_{\phi}(\mathbf{z}_i\vert \mathbf{x}) = \mathcal{N}(\boldsymbol{\mu}_i(\mathbf{x}),\boldsymbol{\Sigma}_i(\mathbf{x})), \\
p_{\theta}(\mathbf{x},\mathbf{z})&=  p_{\theta}(\mathbf{x}\vert \mathbf{z} )p(\mathbf{z}), p_{\theta}(\mathbf{x}\vert \mathbf{z} ) =\mathcal{N}(g_{\theta}(\mathbf{z}_1,\ldots,\mathbf{z}_L) ,\sigma^2), \\
\end{split}
\]
where $g_{\theta}(\mathbf{z}_1,\ldots,\mathbf{z}_L)$ is a deterministic function. The variational principle provides a tractable evidence lower bound (ELBO) by combining Eqs. (\ref{likelihood}) and (\ref{kl}):
\vspace{-2mm}
\begin{equation}
\vspace{-2mm}
\begin{split}
   \mathcal{L}(\theta,\phi,\mathbf{x})&=\log(p_{\theta}(\mathbf{x}))-KL(q_{\phi}(\mathbf{z} \vert \mathbf{x})\Vert p(\mathbf{z} \vert \mathbf{x})) \\
&= E_{q_{\phi}(\mathbf{z}\vert \mathbf{x})}[\log p_{\theta}(\mathbf{x}\vert \mathbf{z})]-KL(q_{\phi}(\mathbf{z} \vert \mathbf{x})\Vert p(\mathbf{z})), 
\end{split}
\end{equation}
which is estimated by reparameterization tricks and sampling ~\cite{DBLP:journals/corr/KingmaW13} and is written in a closed form:
\vspace{-2mm}
\begin{equation}
\begin{split}
    \hat{\mathcal{L}}(\theta,\phi,\mathbf{x}) = \frac{1}{K}\sum_{k=1}^K -\frac{\Vert \mathbf{x}-g_{\theta}(\mathbf{z}_1^k,\ldots,\mathbf{z}_L^k)\Vert_2^2}{\sigma^2}+\\
    \sum_{l=1}^L[\Vert \boldsymbol{\mu}_l\Vert_2^2+trace(\boldsymbol{\Sigma}_l)-\log \vert \boldsymbol{\Sigma}_l\vert].
\end{split}
\vspace{-2mm}
\end{equation}

\iffalse
We calculate $\mu_i(x)$,$\Sigma_i (x)$ and $ g_{\theta}(z_1,\ldots,z_L)$ in an iterative way :
\[
\begin{split}
f_0 &= x\\
f_l &= I_l(f_{l-1})\\
\mu_{l}(x)&=U_l f_l\\
\Sigma_{l}(x)&=S_l f_l,l=1,\ldots,L\\
h_{L} &= z_{L}\\
h_{l}&=G_l(h_{l+1})+C_l z_l,l=1,\ldots,L-1\\
g_{\theta}(z_1,\ldots,z_L)  &= h_1
\end{split}
\]
where the transform $I_l,G_l$ represent dense block in inference process and U-Net decode block in generative process and  $C_l,U_l,S_l$ are matrices.
\fi

Specifically, we take the learned posterior distribution $q_{\phi}(\mathbf{z} \vert \mathbf{x})$ of latent variables from the inference model as the representation.

%Experiments show that our DUHiV achieves  significantly lower ELBO and generates much better reconstruction results than DLGM \cite{DBLP:conf/icml/RezendeMW14} and Ladder VAE \cite{DBLP:conf/nips/SonderbyRMSW16} with MLPs.  

%{\color{red}Figure~\ref{} illustrates the architecture of our DUHiV. (unfinished)}

%\noindent \textbf{Hierarchical VAE.}
%Although the dimension of $Z$ is much less than $\mathcal{I}$, its dimension still adds difficulty to apply interactive visual analysis techniques to study and adjust the representations. Therefore, we introduce a hierarchical VAE that constructs a hierarchy of latent variables so that they can be studied in a well-organized manner.

%1.netweork strcuture: encoder densenet and decoder u-net
%2.objective function

%2.通过可视化方法定义pattern的表达
\subsection{Interactive Insight Identification}
\label{minterface}
We perform arithmetic operations on representations of PGPMs derived from DUHiV and develop visualization-based approaches to analyze the semantic changes of the reconstructed PGPM brought by such operations. In this way, we interactively identify an insight set $A=\{a_1,a_2,...,a_K\}$ and generate the representation for each insight in it, denoted as $P=\{\mathbf{p}_1,\mathbf{p}_2,...,\mathbf{p}_K\}$ where $\mathbf{p}_i$ corresponds to the representation of insight $a_i$. Specifically, we address the following tasks which is beyond the capability of automatic methods:

%After the representation $p(z|I_i)$ for each PGPM $I_i$ is derived from DUHiV, our goal is to identify a TSA insight set $A=\{a_1,a_2,...,a_K\}$ where each insight $a_i \in A$ is denoted as a representation vector sampled from the same latent space $Z$. This requires in-depth comprehension of the latent space structure which is beyond the capability of automatic methods. As a result, we introduce visualization-based approaches to address the following tasks:

\noindent \textbf{T1. Understand the landscape of the latent space.}
For a reasonable representation, variations of different latent variables are expected to result in different semantic changes of the reconstructed PGPM. We allow users to walk in the latent space by flexibly adjusting all latent variables and observe the reconstructed PGPM. This helps users to decide which latent variables need to be adjusted to define an appropriate representation of a TSA insight.

\begin{figure*}[!htbp]
\centering
\setlength{\abovecaptionskip}{3pt} 
\includegraphics[width = 1.6 \columnwidth]{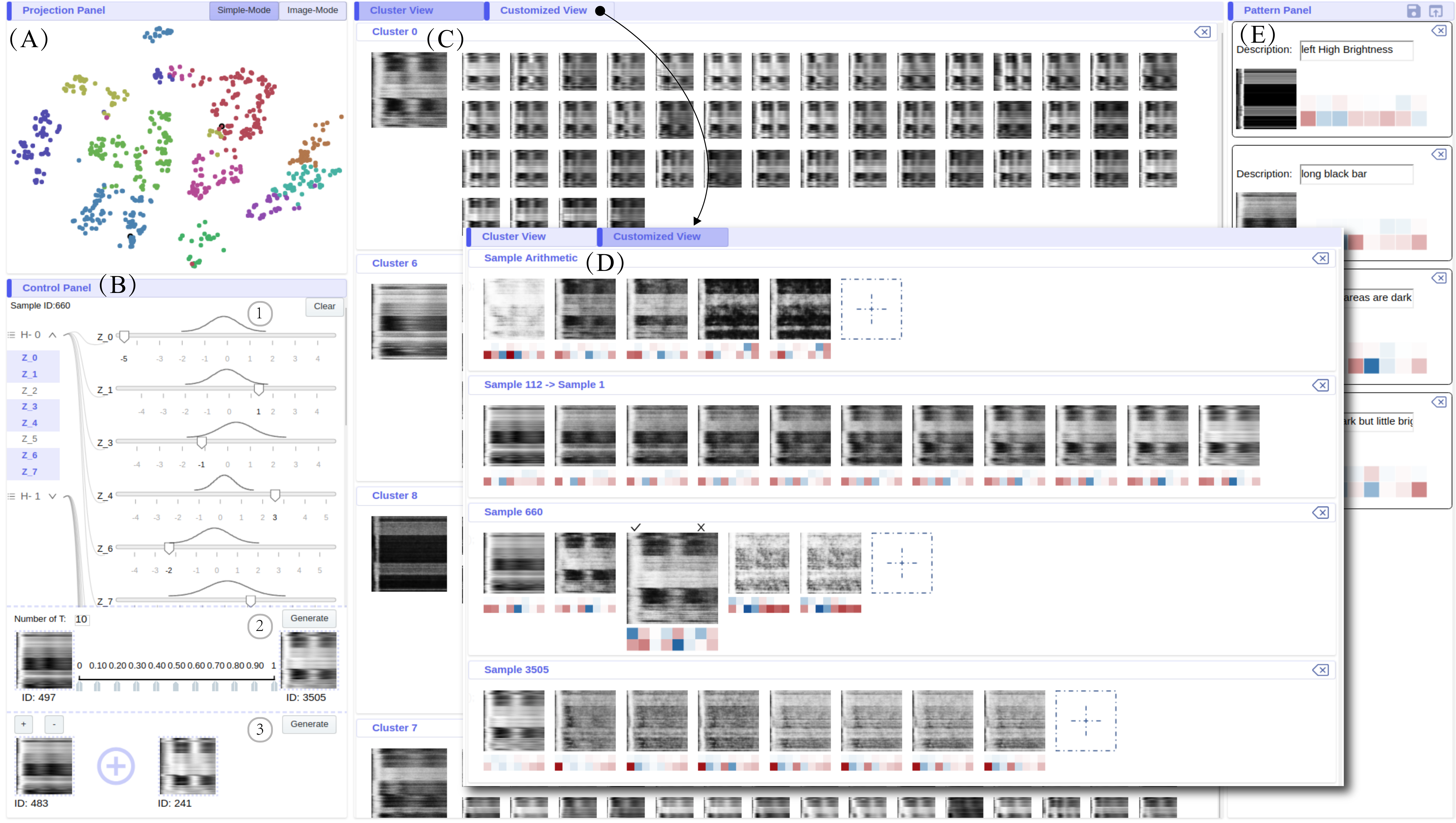}
\caption{The visual interface used to present and analyze the representation of TSA Insights. (A) The projection view that projects PGPMs to a two-dimensional space. (B) The control panel supports arithmetic operations. (C) The cluster view shows original PGPMs for the selected clusters. (D) The analysis view shows semantic changes brought by arithmetic operations on representations. (E) The insight view records the identified insights.}
\label{interface}
\vspace{-6mm}
\end{figure*}

\noindent \textbf{T2. Generate the representation $\mathbf{p}_i$ for a TSA insight $a_i$.}
We provide an efficient definition strategy based on arithmetic operations between representation vectors. Three types of operations are supported. The first operation is the most direct and intuitive way. Given a PGPM $I_i$ and its representation $p(\mathbf{z}|I_i)$, the value of each latent variable $z_i$ is randomly sampled and flexibly adjusted (Figure~\ref{interface} (B1)). The second operation is to perform linear interpolation between the representations of a given PGPMs pair $(I_i,I_j)$ (Figure~\ref{interface} (B2)). The last one is to perform addition and subtraction operations to existing representations (Figure~\ref{interface} (B3)), which is the most efficient way among all operations.

%Compared to semi-supervised methods that use $A$ to label $\mathcal{I}$, defining pattern representation $p(z|a_i)$ directly distinguishes PGPMs with or without pattern $a_i$.

\noindent \textbf{The Visual Interface.}
Figure~\ref{interface} illustrates the visualization interface, consisting of five components (A-E). \textit{The projection view (A)} is the entrance of the interface, in which PGPMs $\mathcal{I}$ is projected to a two-dimensional space using t-distributed Stochastic Neighbor Embedding (t-SNE) and clustered using Kmeans clustering. We use 2-Wasserstein distance to compute the distance between two distributions $I_i\sim \mathcal{N}(\boldsymbol{\mu}_i,\boldsymbol{\Sigma}_i)$ and $I_j\sim \mathcal{N}(\boldsymbol{\mu_j},\boldsymbol{\Sigma}_j)$:
\vspace{-3mm}
\begin{equation}
 W_2(p_i,p_j)=\|\boldsymbol{\mu}_i- \boldsymbol{\mu}_j\|^2_2+tr(\boldsymbol{\Sigma}_i+\boldsymbol{\Sigma}_j-2(\boldsymbol{\Sigma}^{1/2}_j\boldsymbol{\Sigma}_i\boldsymbol{\Sigma}^{1/2}_j)^{1/2}).   
\end{equation}
To avoid visual occlusion and preserve the overall distribution of $\mathcal{I}$ as much as possible, we exploit blue noise sampling on $\mathcal{I}$ and only project the sampled ones.
%We also display the PGPM generated by the average distribution of each cluster (figure). 
Users start interacting with the interface by selecting clusters of interest and displaying them in \textit{the cluster view (C)}. For a selected cluster, we display the PGPMs on the right and display the average PGPM of this cluster on the left. The average PGPM is generated by reconstructing from the average representation. Then users select PGPMs in the cluster view and perform the three type of arithmetic operations in \textit{the control panel (B)}, as mentioned in \textbf{T2}. Specifically, interpolation between $I_i\sim \mathcal{N}(\boldsymbol{\mu}_i,\boldsymbol{\Sigma}_i)$ and $I_j\sim \mathcal{N}(\boldsymbol{\mu}_j,\boldsymbol{\Sigma}_j)$ is computed as:
\vspace{-2mm}
\begin{equation}
\vspace{-2mm}
 I_t\sim \mathcal{N}(t\boldsymbol{\mu}_i+(1-t)\boldsymbol{\mu}_j,(t\boldsymbol{\Sigma}^{1/2}_i+(1-t)\boldsymbol{\Sigma}^{1/2}_j)^2),   
\end{equation}
where $t\in [0,1]$ and $I_t$ equals to $I_i$ and $I_j$ when t=1 and 0, respectively. The interpolation parameters can be flexibly adjusted in the interface. After arithmetic operations, we generate representation vectors and display them in \textit{the analysis view (D)} together with the corresponding reconstructed PGPM. The representation vector is displayed by heatmap, in which the color of each block indicates the sample value of a latent variable, with bluer color indicates lower value and redder indicates higher value. When a generated representation is selected to represent a TSA insight, it is recorded in \textit{the insight view (E)}. Its heatmap and reconstructed PGPM are displayed. Users can also add descriptions to it.
%The APGPM is computed as:
%$$\mathcal{N}(\frac{\Sigma\mu_i}{n},(\frac{\Sigma\Sigma^{1/2}_i}{n})^2)$$,
%where $p_i(z|I_i)\sim \mathcal{N}(\mu_i,\Sigma_i)$ is the distribution of PGPMs and $n$ is the number of PGPMs in the cluster. 
%Since latent variables are organized in a hierarchical structure in which each layer corresponds to features of different semantic levels, the interactive adjustment process is always well-organized and efficient. 

\noindent \textbf{Generating an Insight Representation.}
With the support of the visual interface, defining an insight representation is done in three steps.

First, we need to identify an interested insight $a_i$. The visual interface provides two efficient ways to do this. One is to observe the PGPMs of a selected cluster and discover an insight existing in most of them. The other is to observe outliers of a cluster and analyze whether a special insight exists. 
%In this stage, users discover a set of representative PGPMs $\mathcal{M}$, each containing multiple patterns.

Second, we need to define the representation for $a_i$. We can generate representations by using the three type of arithmetic operations and observe changes in the reconstructed PGPM. The $\mathbf{p}_i$ is determined when its reconstructed PGPM contains and only contains $a_i$.

Finally, we record the representation $\mathbf{p}_i$ and add descriptions to it. By iteratively repeating this process, we finally obtain an insight set $A=\{a_1,a_2,...,a_K\}$ and its corresponding representation set $P=\{\mathbf{p}_1,\mathbf{p}_2,...,\mathbf{p}_K\}$.

%3.根据2中定义出的pattern表达，在PGPM中识别pattern
\subsection{Insight Annotation}
\label{mannotation}
Based on the outputs of Section~\ref{mmodel} and Section~\ref{minterface}, we propose two methods to annotate the dataset $\mathcal{I}$ with insight labels in $A$.

%So far, we have obtained the representation $p(z|I_i)$ of each PGPM $I_i \in \mathcal{I}$ and a TSA insight set $A=\{a_1,a_2,...,a_K\}$. Now we introduce two modes to annotate the dataset $\mathcal{I}$ with $A$ : the automatic mode and the interactive mode.

\noindent \textbf{Semi-supervised Method}
In the semi-supervised way, we identify insights $A$ but do not necessarily generate representations $P$ for them. Instead, users manually annotate part of the dataset directly with the identified insight categories during the exploration process. The rest of the dataset is then automatically annotated in a semi-supervised way in which the DUHiV model learns from the unlabeled data and classifies on the labeled data.

\noindent \textbf{Unsupervised Method}
In the unsupervised way, we use the previously identified insight set $A$ and the generated representation set $P$. We annotate the dataset $\mathcal{I}$ by computing the similarity between each $\mathbf{p}_i$ and each PGPM $I_i$. In our experiment, we compute the Euclidean distance between $\mathbf{p}_i$ and the representation vector of $I_i$, which is sampled by taking the mean of every marginal distribution in $q_{\phi}(\mathbf{z} \vert \mathbf{I})$. Therefore, for each $I_i$, we yield a similarity vector $s_i=[s_i^1,...,s_i^K]$, where $s_i^j$ representing the probability of annotating $I_i$ with insight $a_j$ and higher $s_i^j$ indicates higher probability.

\section{Evaluation}

We conduct three experiments on two real-world PGPM datasets. The first demonstrates the superiority of the DUHiV architecture over other hierarchical VAEs. The second presents qualitative case studies to show the effectiveness of the identification process in our visual interface. The last evaluates the annotation accuracy of our framework. We first introduce our test datasets and then explain our experiments.

\subsection{Data Description}
\label{dataset}
We evaluate our method on two real-world PGPM datasets. The first dataset contains 3,504 PGPMs, denoted as PGPM-3K. The size of each PGPM is $368 \times 464$, from a small-size power grid in Northern China that contains 368 buses and is simulated at 464 time points. The time interval between adjacent time points is 0.01 second. The grayscale of pixel $(i,j)$ depicts the \textit{voltage} value of bus $i$ at time $j$. The second dataset is in the same form with the first one. It contains 250,925 PGPMs, denoted as PGPM-250k, whose size is $800 \times 640$. The time interval between adjacent time points is 0.03 second.

Specifically, PGPM-3K is manually labeled with 10 labels in cooperation with domain experts from China Electric Power Research Institute (CEPRI). On average, each PGPM is annotated with approximately 2.5 labels. Considering it is extremely time-consuming to label PGPM-250k, quantitative experiments are only performed on PGPM-3K.

\subsection{Sample Generation Performance}
\label{othervae}
We conduct the first experiment on PGPM-3Kn and compare DUHiV with two hierarchical VAE models with MLP architectures, that is, Deep Latent Gaussian Models~\cite{DBLP:conf/icml/RezendeMW14} (DLGM) and Ladder VAE~\cite{DBLP:conf/nips/SonderbyRMSW16} (LVAE). Specifically, we compare samples of generated PGPMs and the evidence lower bound (ELBO) which is widely used VAE models~\cite{DBLP:journals/corr/KingmaW13}.

\noindent \textbf{Experimental Settings.}
For fair comparison, we fine-tune the parameters of all models to achieve best performance on our dataset. For DUHiV, we choose design parameters via empirical validations. Specifically, we use DenseNet-121~\cite{DBLP:conf/cvpr/HuangLMW17} as the structure of the inference net and use a hierarchy of two layers of latent variables of sizes 8 and 8. The model is trained for 1200 epochs using Stochastic gradient descent (SGD) with momentum 0.9 and the batch size is 144. The initial learning rate is set to 0.005. The decay of learning rate is set to 0.9 every 10 epoch after the first 800 epochs and is set to 0.9 every epoch after the next 200 epochs. We also use the warm-up strategy in~\cite{DBLP:conf/nips/SonderbyRMSW16} during the first 300 epochs of training. For DLGM, we use a model consisting of two deterministic layers of 400 hidden units and two stochastic layers of 8 latent variables as used in the NORB object recognition dataset. For LVAE, the implemented model consists of two MLPs of size 512 and 256, and two stochastic layers of 8 latent variables.
%Our code is available at: https://github.com/FengHZ/DUHiV.

\begin{figure*}[!ht]
\centering
\setlength{\abovecaptionskip}{3pt}
\includegraphics[width = 1.7 \columnwidth]{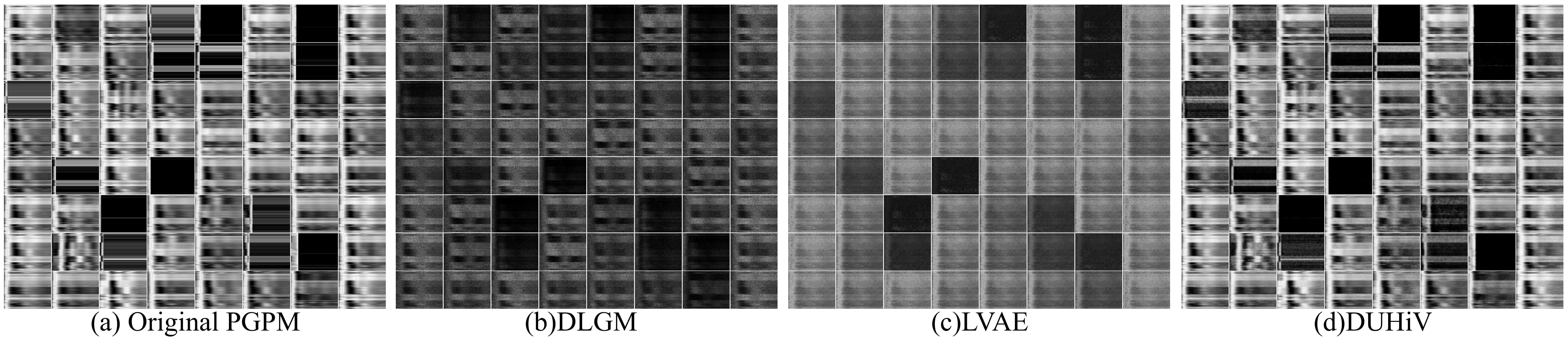}
\caption{samples generated from PGPM-3K by DLGM (b), LVAE (c) and DUHiV (d).}
\label{vaeimage}
\vspace{-5mm}
\end{figure*}

\noindent \textbf{Results.}
We present samples generated by the three models from the original PGPMs in Figure~\ref{vaeimage}. As shown, LVAE captures hue features but filtered out contour information, while DLGM only captures shape features. Instead, DUHiV captures more comprehensive features and therefore achieves the best generation result. The evidence lower bound (ELBO) on the train and test sets in Table~\ref{ELBO} also shows significantly performance improvement of DUHiV over DLGM and LVAE. All these improvements indicates the superiority of our DenseU architecture than the commonly-used MLPs structure over two aspects. First, U-Net extracts features of different levels, including low-level visual patterns and high-level semantic correspondence. Second, the dense block ensures the difference of these features. Therefore, the higher-quality PGPMs generated by DUHiV can further enhance the efficiency and effectiveness of insight identification in our visual interface.

%We also compare the evidence lower bound (ELBO) in Table~\ref{ELBO}. The superior performance improvement of DUHiV over other baselines indicates the benefits of our DenseU architecture. 

%The clearer images reconstructed by DUHiV help users define more effective representations of TSA insights in the interactive mode and therefore better annotation result is achieved. 

\begin{table}
\centering
\setlength{\abovecaptionskip}{3pt}
%\captionsetup{labelformat=empty}
%\caption{asdbdsa}
\begin{tabular}{c|c|c} 
\hline
\multirow{2}{*}{Model} & \multicolumn{2}{c}{$\leq log(p(x))$}  \\ 
\cline{2-3}
                       &Train &Test          \\ 
\hline
\hline
DLGM                   &-25627.5       &-25520.0              \\
LVAE                   &-13630.9       &-13624.3               \\ 
\hline
DUHiV                  &\textbf{-2345.4}       &\textbf{-2416.5}             \\
\hline
\end{tabular}
\caption{Comparison results of evidence lower bound (ELBO) of our method and the representative VAE models on PGPM-3K. The best results are marked in bold.}
\label{ELBO}
\vspace{-6mm}
\end{table}

\subsection{Insigt Identification}
We conduct the second experiment on both PGPM-3K and PGPM-250K to demonstrate the effectiveness of our method in TSA insight identification. 

\begin{figure*}[!ht]
\centering
\setlength{\abovecaptionskip}{3pt}
\includegraphics[width = 1.7 \columnwidth]{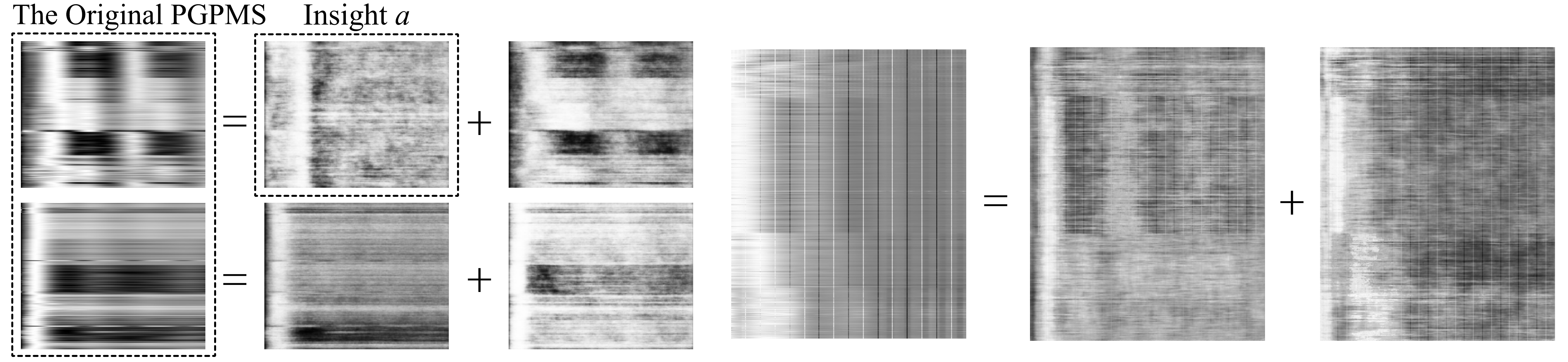}
\caption{Reconstructed PGPMs of TSA insights. \textbf{Left:} PGPM-3K. \textbf{Right:} PGPM-250K.}
\label{experiment3}
\vspace{-3mm}
\end{figure*}

Figure~\ref{experiment3} shows example PGPMs reconstructed from the TSA insight representations. By interacting with our visual interface, we can decompose an original PGPM into a combination of several insights, generate a representation for each insight, and use the representation to reconstruct the PGPM. The reconstructed PGPM are inevitably blurry to some extent because it is not necessarily in the generation space of the dataset. Despite the blur, we can notice that the reconstructed PGPM successfully captures the main features of an insight.

Take insight $a$ in Figure~\ref{experiment3} as an example, we explain how we define its representation with the support of the visual interface. We use the two-layer hierarchy of latent variables of sizes 8 and 8. Latent variables in the second layer determine the overall insight and the semantic correspondence, and the ones in the first layer determine detailed local shape. So we decide to first adjust the second layer to obtain the target insight and then adjust the first layer for local optimization. Particularly, different latent variables in the second layer determine different aspects of the generated PGPM: the hue, the horizontal/vertical position, the width/height of the insight, etc.. As shown, insight $a$ represents the sudden increase of voltage shortly after the start time. An intuitive way to define the representation of insight $a$ is to start from the corresponding original PGPM and remove the four black blocks. Therefore, we first adjust the latent variable controlling the hue in the second layer, then adjust the first layer to form a clearer shape and finally obtain an effective representation.

\subsection{Insight Annotation}
We conduct the last experiment on PGPM-3K. We apply DUHiV as a feature extractor and compare it with other three unsupervised representation learning models: Non-Parametric Instance Discrimination~\cite{DBLP:conf/cvpr/WuXYL18} (NPID) which is the state-of-the-art on ImageNet classification, and DLGM, LVAE in Section~\ref{othervae}. We demonstrate the quality of the learned features by evaluating the annotation performance using these features. Specifically, we use average precision (AP) and mean average precision (mAP), which are widely used in multi-label learning algorithms~\cite{DBLP:conf/aaai/LiHJZ12}. For fair comparison, We use the same DenseNet-121\cite{DBLP:conf/cvpr/HuangLMW17} as the CNN backbone for NPID. Other experimental settings are the same with Section~\ref{othervae}.

\begin{figure}[!ht]
\centering
\setlength{\abovecaptionskip}{3pt}
\includegraphics[width = 0.85 \columnwidth]{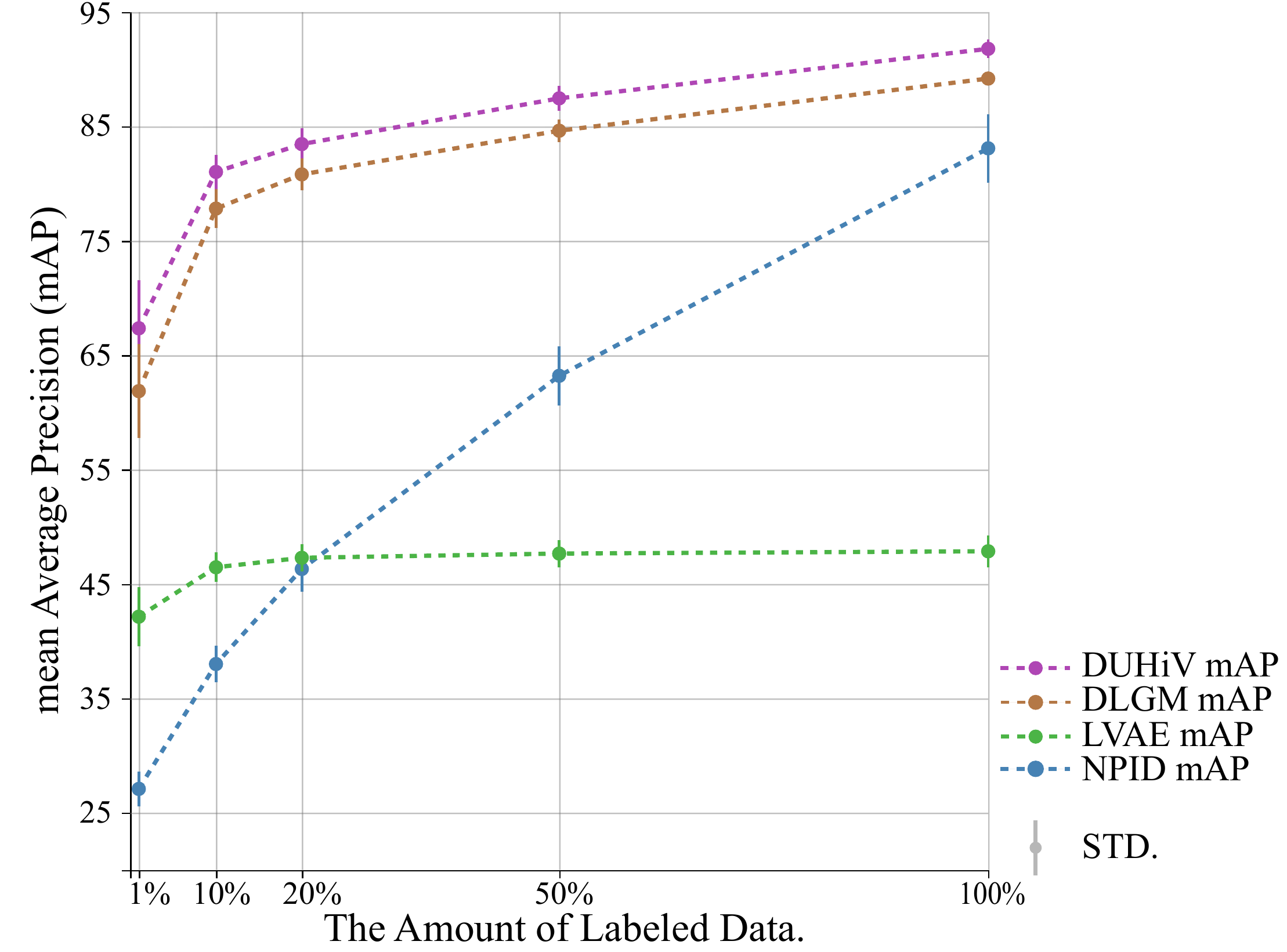}
\caption{Comparison Results of mean Accuracy (mA) and mean Average Precision (mAP) in automatic mode with an increasing fraction of labeled data (x axis).}
\label{semi}
\vspace{-5mm}
\end{figure}

\noindent \textbf{Semi-supervised Method.}
%For fair comparison, we build KNN classifiers using the learned representations for all models to annotate the dataset. We perform 5-fold cross validation on PGPM-3K. The proportion of the labeled dataset varies from $1\%$ to $20\%$ of the entire dataset. Figure~\ref{semi} shows that our method outperforms all baseline models. When only $1\%$ of the dataset is labeled, we achieve $86.1\%$ mean accuracy and $67.4\%$ mAP, demonstrating the effectiveness of our learned features.
For fair comparison, we build KNN classifiers using the learned representations for all models to annotate the dataset. We perform 5-fold cross validation on PGPM-3K. The proportion of the labeled dataset varies from $1\%$ to $100\%$ of the entire dataset. Figure~\ref{semi} shows that DUHiV and DLGM outperform LVAE and NPID by a large margin. The trivial performance of LVAE is because it captures shape features only (Section~\ref{othervae}). Therefore, DLGM can achieve better mAP with tighter ELBO than LVAE. We also notice that when $10\%$ of the dataset is labeled, DUHiV can achieve a $81.03\%$ mAP, which is only $10.7\%$ less than the $100\%$-labeled situation. This advantage can significantly reduce annotation cost and demonstrates superiority over NPID which is sensitive to the annotation proportion.

\noindent \textbf{Unsupervised Method.}
%The interactive mode works in an unsupervised way, in which all models are used as a feature extractor. By performing arithmetic operations of the learned representations in our visual interface, we obtain the representation of TSA insights and annotate the dataset. 
For fair comparison, we interactively identify insights by using DUHiV and use all models to learn the features of the reconstructed PGPMs of the identified insights. 
%We use Euclidean distance to measure the similarity between representation vectors. 
We report the 5-fold cross validation results in Table~\ref{accuracy}. As shown, our method achieves the best mAP of $64.4\%$, leading to a nearly $38.7\%$ performance improvement over NPID, $34.1\%$ improvement over LVAE and $15.4\%$ improvement over DLGM. To be noticed, the unsupervised interactive mode enables accurate annotation by achieving comparable mAP to the semi-supervised automatic mode with $1\%$ labeled data. It indicates that with the support of our visual interface, the generated representations of insights are effective enough to approach a ground-truth level.

\begin{table*}
\centering
\setlength{\abovecaptionskip}{3pt}
\setlength{\tabcolsep}{0.7mm}{
\begin{tabular}{ccccccccccc|c} 
\hline
Model       & P1 & P2 & P3 & P4 & P5 & P6 & P7 & P8 & P9 & P10                   & mAP          \\ 
\hline
\hline
DLGM     &\textbf{98.9$\pm$0.3}    &64.9$\pm$4.8    &70.2$\pm$1.7    &33.9$\pm$4.4    &34.9$\pm$2.0    &45.5$\pm$5.9    &32.2$\pm$9.1    &48.6$\pm$2.8    &29.9$\pm$1.8    &31.3$\pm$6.3                         &49.0$\pm$1.1               \\
LVAE         &46.9$\pm$16.5    &20.9$\pm$2.3    &32.7$\pm$7.8    &23.9$\pm$1.1    &62.3$\pm$6.6    &16.6$\pm$2.6    &35.7$\pm$1.7    &19.7$\pm$2.3    &38.6$\pm$2.7    &6.1$\pm$2.4                          &30.3$\pm$2.7               \\
NPID        &26.0$\pm$4.9   &20.1$\pm$2.5    &24.3$\pm$2.0    &22.7$\pm$2.1   &53.0$\pm$2.6    &14.7$\pm$2.4    &32.1$\pm$3.5    &17.7$\pm$2.4    &35.4$\pm$3.8   & 10.9$\pm$7.1                        &25.7$\pm$1.4               \\
\hline
our method  &72.5$\pm$3.9    &\textbf{71.0$\pm$5.0}    &\textbf{77.3$\pm$2.5}    &\textbf{38.4$\pm$7.4}    &\textbf{80.7$\pm$3.5}   &\textbf{50.8$\pm$6.3}    &\textbf{73.2$\pm$7.2}    &\textbf{64.8$\pm$7.3}    &\textbf{49.7$\pm$5.9}    &\textbf{66.2$\pm$6.9}                    &\textbf{64.4$\pm$1.1}              \\
\hline
\end{tabular}}
%\captionsetup{labelformat=empty}
\caption{Comparison results of AP and mAP in $\%$ of our method and the baselines on the PGPM-3K dataset (mean $\pm$ std.). Column P$i$ indicates the category of insight $i$. The best results are marked in bold.}
\label{accuracy}
\vspace{-5mm}
\end{table*}

\section{Related Work}
%The literature has shown abundant image pattern mining models~\cite{boutell2002review}, including supervised models represented by convolutional neural networks (CNN)~\cite{krizhevsky2012imagenet} and unsupervised models like clustering~\cite{jain1988algorithms,xie2016unsupervised}. Although these models have show impressing performance on natural scenes, they are barely studied on chart images, which is usually perceived in a different way~\cite{cleveland1984graphical}. 

\textbf{Automatic Insight Identification.}
Insights, in the sense of knowledge discovery and data mining, are interesting facts underly the data~\cite{lin2018bigin4}. To accelerate and simplify the tedious insight identification process, KDD experts achieve automatic insight identification by storing data in a specified data structure, for example, a data cube, and applying approximate query processing techniques to speed up the query~\cite{lin2018bigin4,tang2017extracting}. In this paper, we extend this concept to data in the form of chart images and adopt deep learning techniques to achieve efficient interactive insight identification.

\textbf{Chart Image Recognition.}
Existing chart image recognition mainly focuses on two tasks: chart type classification and visual contents decoding~\cite{mishchenko2011chart}. Insight identification shares a common framework with these two tasks that starts from feature extraction and ends at insight extraction. For the feature extraction part, prior models use hand-crafted features~\cite{savva2011revision} but does not scale well with large amount of unbalanced data. Amara et. al.~\cite{amara2017convolutional} leverage convolutional neural network (CNN) on image sets and achieve better classification results. Different from these methods, our method supports unsupervised feature extraction and arithmetic feature operations to generate higher-quality features. For the insight extraction part, existing works mainly rely on supervised classification or unsupervised clustering which output label numbers instead of semantic insights. In contrast, we adopt visualization-based approaches to interactively and directly define and extract insights based on the learned representations.

\textbf{Disentangled Representation Learning.} Disentanglement requires different independent variables in the learned representation to capture independent factors that generate the input data. Early methods are based on denoising autoencoders~\cite{DBLP:journals/corr/KingmaW13} and restricted Boltzmann machines~\cite{hinton2006fast}. However, deep generative models, represented by variational autoencoder (VAE)~\cite{maaloe2015improving} and generative adversarial net (GAN)~\cite{goodfellow2014generative}, have recently achieved better results in this area due to their ability in preserving all factors of variation~\cite{DBLP:journals/corr/abs-1812-05069}. Mathieu et. al.~\cite{mathieu2016disentangling} introduce a conditional generative model that leverages both VAE and GAN. Their model learns to separate the factors related to labels from another source of variability based on weak assumptions. InfoGAN~\cite{chen2016infogan} makes a further progress by training without any kind of supervision. Being an information-theoretic extension to GAN, it maximizes the mutual information between subsets of latent variables and the observation. The main drawback of these approaches is the lack of interpretability between latent variables and aspects of the generated image. In contrast, we follow
~\cite{DBLP:conf/icml/RezendeMW14} and ~\cite{DBLP:conf/nips/SonderbyRMSW16} to build a novel hierarchical VAE architecture to ease this problem by building hierarchies of latent variables. 
%(Add introductions to hierarchical vae if necessary.)

%要不要说为啥用vae不用别的representation learning的方法？？
%1.vae可以用来学习表征，怎么学的（简述原理），学到的是啥
%2.普通的vae学到的为啥不行，为啥需要用Hvae，hvae学到的是啥
%3.Hvae models介绍
%\textbf{Variational Autoencoder Based Representation Learning.}
%We adopt variational Autoencoder (VAE) to parametrically learn the representation~\cite{kingma2013auto,tschannen2018recent}, which turns out to be the posterior distribution $p_{\theta}(z|x)$ on the latent space given an observed input $x$. Models such as $\beta$-VAE~\cite{higgins2016beta}, FactorVAE~\cite{kim2018disentangling} and InfoVAE~\cite{zhao2017infovae} disentangle $x$ so that different variables $z_i$ in $p_{\theta}(z|x)$ capture independent factors that are assumed to generate $x$. This assumption makes it possible to define semantic patterns (a combination of factors) by adjusting $z$ in $p_{\theta}(z|x)$. However, analysts may still get confused with which $z_i$ to be adjusted since the size and sort of $z$ hardly reach trial level. Hierarchical VAE solves this problem by constructing a hierarchy of $z$ so that $z_i$ in $z$ can be studied in a grouped manner~\cite{gulrajani2016pixelvae,sonderby2016ladder,van2017neural}. (Add introductions to hierarchical vae if necessary.)

\section{Conclusion}
In this paper, we propose an interactive insight identification and annotation framework for transient stability insight discovery in power grid pixel maps. We develop a DenseU-hierarchical variational autoencoder combined with interactive visualization-based approaches for representation learning of transient stability insights. To the best of our knowledge, this is the first work on interactive insight identification and annotation in power grid images and also on learning more refined representations of insights. Experiments using real-world datasets indicate the improvement of our method compared to baselines.

\begin{spacing}{0.9} 
%%行间距变为double-space  
\bibliographystyle{named}
\bibliography{ijcai19-multiauthor}
\end{spacing}

\end{document}